\documentclass{article}

\PassOptionsToPackage{numbers,sort&compress}{natbib}

\usepackage[eandd, final]{neurips_2026}

\usepackage[utf8]{inputenc} % allow utf-8 input
\usepackage[T1]{fontenc}    % use 8-bit T1 fonts
\usepackage{hyperref}       % hyperlinks
\usepackage{url}            % simple URL typesetting
\usepackage{booktabs}       % professional-quality tables
\usepackage{amsfonts}       % blackboard math symbols
\usepackage{nicefrac}       % compact symbols for 1/2, etc.
\usepackage{microtype}      % microtypography
\usepackage{xcolor}         % colors
\usepackage{graphicx} 
\usepackage{multirow}
\usepackage{colortbl}
\usepackage{makecell}
\usepackage{amssymb}
\usepackage{tikz}
\usepackage{float}
\usepackage{subcaption}
\usepackage{array}
\usepackage[table]{xcolor}
\usepackage{enumitem}
\usepackage{amsmath}

\usepackage{titletoc}
\usepackage{etoc}
\newcommand{\opensourcemasks}{%
\begin{table}[H]
\centering
\footnotesize
\setlength{\tabcolsep}{6pt}
\renewcommand{\arraystretch}{1.2}
\begin{minipage}{0.22\linewidth}
    \centering
    \includegraphics[width=\linewidth]{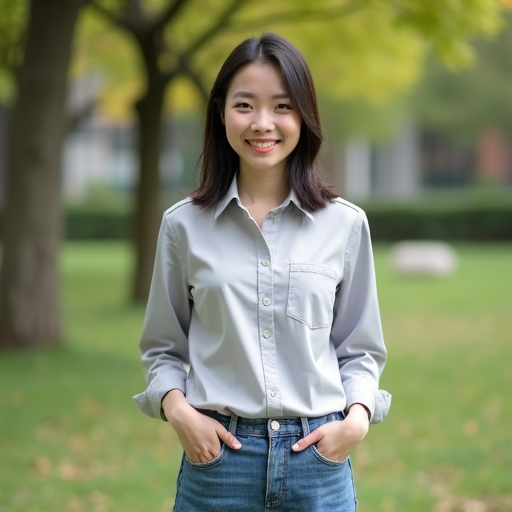}
    \subcaption{Input image}\label{subfig:input}
\end{minipage}%
\hfill
\begin{minipage}{0.76\linewidth}
\begin{tabular}{@{} c p{0.85\linewidth} @{}}
\toprule
\textbf{Pipeline} & \textbf{Masking details} \\
\midrule
1 &
\begin{itemize}[leftmargin=*, nosep, topsep=2pt, partopsep=0pt]
    \item GroundingDINO + SAM\_HQ
    \item prompts: ``shirt'', ``clothes'', ``trousers''
    \item Dilate masks by 8px
\end{itemize} \\

2 &
\begin{itemize}[leftmargin=*, nosep, topsep=2pt, partopsep=0pt]
    \item Segformer\_b3\_fashion, Segformer\_b2\_clothes
    \item Dilate mask by (14,7)
\end{itemize} \\

3 &
\begin{itemize}[leftmargin=*, nosep, topsep=2pt, partopsep=0pt]
    \item GroundingDINO + SAM
    \item prompts: ``woman'', ``dress'', ``lingerie'', ``underwear'', ``swimsuit'', ``pants'', etc.
    \item combine segmentation masks with overlay
\end{itemize} \\

4 &
\begin{itemize}[leftmargin=*, nosep, topsep=2pt, partopsep=0pt]
    \item Hand-drawn mask
\end{itemize} \\

5 &
\begin{itemize}[leftmargin=*, nosep, topsep=2pt, partopsep=0pt]
    \item SegBody with face=FALSE
\end{itemize} \\

6 &
\begin{itemize}[leftmargin=*, nosep, topsep=2pt, partopsep=0pt]
    \item Mirrors IDM-VITON (ECCV 2024)
    \item SCHP LIP + ATR checkpoints, with Densepose + Openpose
\end{itemize} \\
\bottomrule
\end{tabular}
\subcaption{Masking steps for nudification and un/redressing pipelines}\label{subtab:pipelines}
\end{minipage}
\caption{Masks used by real-world nudification and redressing pipelines vary substantially in shape, extent, and the body regions they cover. The pipelines surveyed include both those that explicitly advertise nudification, and general-purpose redressing pipelines (which follow the same technical patterns and are equally capable of producing nude outputs). Because a defender cannot anticipate which mask the attacker will use, defenses optimized against a specific protection mask transfer poorly. All pipelines surveyed are open-source. We do not cite them to avoid amplification.}
\label{tab:nudificationpipeline-mask}
\end{table}
}

\newcommand{\protectionoverviewnewshort}{

\begin{table*}[t]
\centering
\small
\renewcommand{\arraystretch}{0.6}
\setlength{\tabcolsep}{5pt}
\begin{tabular}{l l p{2cm} p{2cm} p{2.5cm} p{2cm}}
\toprule
Variant & Attack & Sexualization (max=2) & Naturalness & Recognizability & DefenseScore (max=1) \\
\midrule
\multicolumn{6}{l}{\textbf{Unprotected}} \\

\multirow{3}{*}{--} & Body & 1.023 & 21.9 & 0.602 & 0.489 \\
 & Body + Context & 1.612 & 22.7 & 0.550 & 0.194 \\
& Dilated Body (d=7) & 1.708 & 23.3 & 0.599 & 0.146 \\
\midrule
\multicolumn{6}{l}{\textbf{PhotoGuard}~\citep{salman2023raising}} \\

\multirow{3}{*}{Body} & \cellcolor{nuditycolor} Body & \cellcolor{nuditycolor} 1.344 & \cellcolor{nuditycolor} 22.2 & \cellcolor{nuditycolor} 0.104 & \cellcolor{nuditycolor} 0.328 \\
 &   Body + Context & 1.620 & 22.3 & 0.193 & 0.190 \\
 & Dilated Body (d=7) & 1.652 & 22.5 & 0.105 & 0.174 \\
\cmidrule(l){1-6}
\multirow{3}{*}{Body + Context} & Body & 1.195 & 22.0 & 0.241 & 0.403 \\
 & \cellcolor{nuditycolor} Body + Context & \cellcolor{nuditycolor} 1.598 & \cellcolor{nuditycolor} 22.2 & \cellcolor{nuditycolor} 0.188 & \cellcolor{nuditycolor} 0.201 \\
& Dilated Body (d=7) & 1.649 & 22.8 & 0.224 & 0.176 \\

\midrule
\multicolumn{6}{l}{\textbf{DiffusionGuard}~\citep{choi2025diffusionguard}} \\

\multirow{3}{*}{Body} & \cellcolor{nuditycolor} Body & \cellcolor{nuditycolor} 1.160 & \cellcolor{nuditycolor} 21.5 & \cellcolor{nuditycolor} 0.093 & \cellcolor{nuditycolor} 0.420 \\
 &  Body + Context & 1.487 & 21.9 & 0.092 & 0.257 \\
 & Dilated Body (d=7) & 1.256 & 21.5 & 0.093 & 0.372 \\
\cmidrule(l){1-6}
\multirow{3}{*}{Body + Context} & Body & 1.228 & 22.1 & 0.145 & 0.386 \\
& \cellcolor{nuditycolor} Body + Context & \cellcolor{nuditycolor} 1.412 & \cellcolor{nuditycolor} 21.8 & \cellcolor{nuditycolor} 0.127 & \cellcolor{nuditycolor} 0.294 \\
 & Dilated Body (d=7) & 1.569 & 22.5 & 0.136 & 0.215 \\
\midrule
\multicolumn{6}{l}{\textbf{AdvPaint}~\cite{jeon2025advpaint}} \\
\multirow{3}{*}{Body} & \cellcolor{nuditycolor} Body & \cellcolor{nuditycolor} 0.098 & \cellcolor{nuditycolor} 19.3 & \cellcolor{nuditycolor} 0.090 & \cellcolor{nuditycolor} 0.951 \\
 & Body + Context & 1.229 & 21.5 & 0.092 & 0.385 \\
 & Dilated Body (d=7) & 0.523 & 19.9 & 0.089 & 0.739 \\
\cmidrule(l){1-6}
\multirow{3}{*}{BBox} & Body & 1.233 & 21.8 & 0.095 & 0.384 \\
 & Body + Context & 1.313 & 21.7 & 0.091 & 0.344 \\
 & Dilated Body (d=7) & 1.362 & 21.7 & 0.094 & 0.319 \\
\midrule
\multicolumn{6}{l}{\textbf{DiffVax}~\citep{ozden2025diffvax}} \\

\multirow{3}{*}{Body} & \cellcolor{threatmismatchcolor} Body & \cellcolor{threatmismatchcolor} 0.718 & \cellcolor{threatmismatchcolor} 21.0 & \cellcolor{threatmismatchcolor} 0.595 & \cellcolor{threatmismatchcolor} 0.641 \\
 & Body + Context & 1.445 & 22.0 & 0.533 & 0.278 \\
 & Dilated Body (d=7) & 1.683 & 23.0 & 0.593 & 0.159 \\
\cmidrule(l){1-6}
\multirow{3}{*}{Context} & \cellcolor{threatmismatchcolor} Body & \cellcolor{threatmismatchcolor} 0.980 & \cellcolor{threatmismatchcolor} 21.7 & \cellcolor{threatmismatchcolor} 0.586 & \cellcolor{threatmismatchcolor} 0.510 \\
 & Body + Context & 1.658 & 22.2 & 0.536 & 0.171 \\
 & Dilated Body (d=7) & 1.711 & 23.2 & 0.587 & 0.145 \\
\cmidrule(l){1-6}
\multirow{3}{*}{Body + Context} & Body & 0.549 &  20.8 & 0.593 & 0.725 \\
& \cellcolor{nuditycolor} Body + Context & \cellcolor{nuditycolor} 1.541 & \cellcolor{nuditycolor} 22.1 & \cellcolor{nuditycolor} 0.537 & \cellcolor{nuditycolor} 0.230 \\
 & Dilated Body (d=7) & 1.709 & 23.3 & 0.597 & 0.145 \\
\bottomrule
\end{tabular}
\caption{Rows are grouped by protection method, protection variant, and attack mask. All values are means over \bb~under the \texttt{naked} inpainting prompt. Rows highlighted in \colorbox{nuditycolor}{blue} indicate the attack mask matches the protection mask \textit{exactly}. Rows highlighted in \colorbox{threatmismatchcolor}{orange} indicate an irreconcilable threat model mismatch. See Section \ref{sec:background_mismatch} for how this threat model mismatch is defined.}
\label{tab:protection-overview-short}
\end{table*}
}

\newcommand{\bigprotections}{
\begin{table}[H]
\centering
\scriptsize
\renewcommand{\arraystretch}{1.0}
\setlength{\tabcolsep}{5pt}
\begin{tabular}{llrrrr}
\toprule
Variant & Attack & Sexualization & Naturalness & Recognizability & DefenseScore \\
\midrule
\multicolumn{6}{l}{\textbf{Unprotected}} \\

\multirow{5}{*}{--} & body & 1.023 & 21.9 & 0.602 & 0.489 \\
 & body + context & 1.612 & 22.7 & 0.550 & 0.194 \\
 & dilated body (d=7) & 1.708 & 23.3 & 0.599 & 0.146 \\
 & dilated body (d=15) & 1.746 & 23.4 & 0.590 & 0.127 \\
 & dilated body (d=31) & 1.780 & 23.4 & 0.555 & 0.110 \\
\midrule
\multicolumn{6}{l}{\textbf{PhotoGuard}~\citep{salman2023raising}} \\
\multirow{5}{*}{body} & body & 1.344 & 22.2 & 0.104 & 0.328 \\
& \cellcolor{nuditycolor} body + context & \cellcolor{nuditycolor} 1.620 & \cellcolor{nuditycolor} 22.3 & \cellcolor{nuditycolor} 0.193 & \cellcolor{nuditycolor} 0.190 \\
 & dilated body (d=7) & 1.652 & 22.5 & 0.105 & 0.174 \\
 & dilated body (d=15) & 1.697 & 22.6 & 0.109 & 0.152 \\
 & dilated body (d=31) & 1.752 & 22.7 & 0.116 & 0.124 \\
\midrule
\multirow{5}{*}{body+context} & body & 1.195 & 22.0 & 0.241 & 0.403 \\
& \cellcolor{nuditycolor} body + context & \cellcolor{nuditycolor} 1.598 & \cellcolor{nuditycolor} 22.2 & \cellcolor{nuditycolor} 0.188 & \cellcolor{nuditycolor} 0.201 \\
 & dilated body (d=7) & 1.649 & 22.8 & 0.224 & 0.176 \\
 & dilated body (d=15) & 1.689 & 22.9 & 0.224 & 0.156 \\
 & dilated body (d=31) & 1.746 & 23.0 & 0.210 & 0.127 \\
\cmidrule(l){1-6}

\multicolumn{6}{l}{\textbf{DiffusionGuard}~\citep{choi2025diffusionguard}} \\

\multirow{5}{*}{body} & \cellcolor{nuditycolor} body & \cellcolor{nuditycolor} 1.160 & \cellcolor{nuditycolor} 21.5 & \cellcolor{nuditycolor} 0.093 & \cellcolor{nuditycolor} 0.420 \\
 & body + context & 1.487 & 21.9 & 0.092 & 0.257 \\
 & dilated body (d=7) & 1.256 & 21.5 & 0.093 & 0.372 \\
 & dilated body (d=15) & 1.379 & 21.8 & 0.093 & 0.310 \\
 & dilated body (d=31) & 1.587 & 22.1 & 0.096 & 0.206 \\
\midrule
\multirow{5}{*}{body+context} & body & 1.228 & 22.1 & 0.145 & 0.386 \\
& \cellcolor{nuditycolor} body + context & \cellcolor{nuditycolor} 1.412 & \cellcolor{nuditycolor} 21.8 & \cellcolor{nuditycolor} 0.127 & \cellcolor{nuditycolor} 0.294 \\
 & dilated body (d=7) & 1.569 & 22.5 & 0.136 & 0.215 \\
 & dilated body (d=15) & 1.669 & 22.7 & 0.136 & 0.165 \\
 & dilated body (d=31) & 1.748 & 22.8 & 0.139 & 0.126 \\
\cmidrule(l){1-6}

\multicolumn{6}{l}{\textbf{AdvPaint}~\citep{jeon2025advpaint}} \\
\multirow{5}{*}{body} & \cellcolor{nuditycolor} body & \cellcolor{nuditycolor} 0.098 & \cellcolor{nuditycolor} 19.3 & \cellcolor{nuditycolor} 0.090 & \cellcolor{nuditycolor} 0.951 \\
 & body + context & 1.229 & 21.5 & 0.092 & 0.385 \\
 & dilated body (d=7) & 0.523 & 19.9 & 0.089 & 0.739 \\
 & dilated body (d=15) & 1.086 & 21.0 & 0.090 & 0.457 \\
 & dilated body (d=31) & 1.434 & 21.6 & 0.090 & 0.283 \\
\midrule
\multirow{5}{*}{bbox} & body & 1.233 & 21.8 & 0.095 & 0.384 \\
 & body + context & 1.313 & 21.7 & 0.091 & 0.344 \\
 & dilated body (d=7) & 1.362 & 21.7 & 0.094 & 0.319 \\
 & dilated body (d=15) & 1.470 & 21.9 & 0.096 & 0.265 \\
 & dilated body (d=31) & 1.601 & 22.1 & 0.099 & 0.200 \\
\cmidrule(l){1-6}

\multicolumn{6}{l}{\textbf{DiffVax}~\citep{ozden2025diffvax}} \\

\multirow{5}{*}{body} & \cellcolor{threatmismatchcolor} body & \cellcolor{threatmismatchcolor} 0.718 & \cellcolor{threatmismatchcolor} 21.0 & \cellcolor{threatmismatchcolor} 0.595 & \cellcolor{threatmismatchcolor} 0.641 \\
 & body + context & 1.445 & 22.0 & 0.533 & 0.278 \\
 & dilated body (d=7) & 1.683 & 23.0 & 0.593 & 0.159 \\
 & dilated body (d=15) & 1.753 & 23.2 & 0.586 & 0.124 \\
 & dilated body (d=31) & 1.794 & 23.3 & 0.546 & 0.103 \\
 \cmidrule(l){1-6}
\multirow{5}{*}{body+context} & body & 0.549 & 20.8 & 0.593 & 0.725 \\
& \cellcolor{nuditycolor} body + context & \cellcolor{nuditycolor} 1.541 & \cellcolor{nuditycolor} 22.1 & \cellcolor{nuditycolor} 0.537 & \cellcolor{nuditycolor} 0.230 \\
 & dilated body (d=7) & 1.709 & 23.3 & 0.597 & 0.145 \\
 & dilated body (d=15) & 1.740 & 23.3 & 0.590 & 0.130 \\
 & dilated body (d=31) & 1.776 & 23.4 & 0.551 & 0.112 \\
\cmidrule(l){1-6}
\multirow{5}{*}{context} & \cellcolor{threatmismatchcolor} body & \cellcolor{threatmismatchcolor} 0.980 & \cellcolor{threatmismatchcolor} 21.7 & \cellcolor{threatmismatchcolor} 0.586 & \cellcolor{threatmismatchcolor} 0.510 \\
 & body + context & 1.658 & 22.2 & 0.536 & 0.171 \\
 & dilated body (d=7) & 1.711 & 23.2 & 0.587 & 0.145 \\
 & dilated body (d=15) & 1.738 & 23.3 & 0.578 & 0.131 \\
 & dilated body (d=31) & 1.785 & 23.3 & 0.542 & 0.107 \\

\bottomrule
\end{tabular}
\caption{Full results across all mask dilations. Rows are grouped by protection method, protection variant, and attack mask. Naturalness is computed with PickScore~\citep{kirstain_pick--pic_nodate}. Recognizability is computed with ArcFace~\citep{deng2019arcface}, measured against the original unattacked source-image face; values range from $-1$ to $1$, with $1$ indicating identical identity. All values are means over the benchmark under the \texttt{naked} inpainting prompt. Rows highlighted in \colorbox{nuditycolor}{blue} match the nudification threat model and indicate where the attack mask matches the protection mask. Rows highlighted in \colorbox{threatmismatchcolor}{orange} indicate an irreconcilable threat-model mismatch between nudification and the method's assumptions.}
\label{tab:protection-overview-full}
\end{table}
}

\definecolor{darkred}{RGB}{139, 0, 0}
\definecolor{nuditycolor}{RGB}{219, 234, 254} 
\definecolor{threatmismatchcolor}{RGB}{254, 230, 206} 
\definecolor{lightred}{RGB}{254, 215, 215}     
\definecolor{paleblush}{RGB}{255, 245, 245}   

\let\titleold\title
\renewcommand{\title}[1]{\titleold{#1}\newcommand{\thetitle}{#1}}
\makeatletter
\newcommand{\maketitlesupplementary}{
  \newpage
  \begin{center}
    \Large
    \textbf{\@title}\\
    \vspace{0.5em}
    Appendix\\
    \vspace{1.0em}
  \end{center}
}
\makeatother

\title{\textsc{BodyBench}: Evaluating Adversarial Image Defenses Against AI Nudification Inpainting}

\newcommand{\bb}{\textsc{BodyBench}}
\newcommand{\diffvax}{\textsc{DiffVax}~}
\newcommand{\photoguard}{\textsc{PhotoGuard~}}
\newcommand{\diffguard}{\textsc{DiffusionGuard~}}
\newcommand{\advpaint}{\textsc{AdvPaint}~}

\author{%
  Li Qiwei\thanks{Equal contribution.} \\
  University of Michigan \\
  \texttt{rrll@umich.edu} \\
  \And
  Salma Abdel Magid\footnotemark[1] \\
  Princeton University \\
  \texttt{sabdelmagid@princeton.edu} \\
  \AND
  Olga Russakovsky \\
  Princeton University \\
  \texttt{olgarus@princeton.edu}
  \And
  Eric Gilbert \\
  University of Michigan \\
  \texttt{eegg@umich.edu}
  \And
  Sarita Schoenebeck \\
  University of Michigan \\
  \texttt{yardi@umich.edu}
}
\begin{document}

\maketitle
\begin{abstract}
Adversarial perturbation defenses protect images from unauthorized edits by adding imperceptible noise. Yet none of these methods have been evaluated against nudification, despite AI-generated non-consensual intimate imagery (AIG-NCII) being among the most documented harms of generative AI. Moreover, nudification edits are readily carried out through inpainting, the primary setting these protection methods are designed to defend against. We argue this is a problem-formulation failure since prior threat models do not correspond to nudification edits (also called \textit{attacks}). The misalignment of existing approaches is structural, spanning mask shape, training and testing prompt distributions, and success criterion which causes existing evaluations to systematically overestimate protection effectiveness. We reformulate image protection against unauthorized editing around this documented harm. We introduce \bb, a benchmark of 648 fully synthetic clothed adult subjects, each paired with 11 mask variants spanning body, face, and context regions. We propose three harm-aligned metrics: \textit{sexualization}, \textit{naturalness}, and \textit{recognizability} that jointly determine whether an inpainted output constitutes a successful nudification attack. Auditing four recent protection methods (PhotoGuard, DiffusionGuard, AdvPaint, DiffVax) on \bb, we find that protection succeeds in only one configuration of one method, and only under exact mask alignment between defender and attacker---a condition no real defender can guarantee and one that we show is easily always bypassed. We also show that standard evaluation metrics such as peak signal-to-noise ratio (PSNR) are essentially uncorrelated with nudification protection. \bb~makes this misalignment measurable and provides a foundation for defenses grounded in documented harm. The code and dataset are publicly available at 
\url{https://github.com/gauchewy/BodyBench/tree/main}.

\end{abstract}

\begin{center}
\textcolor{darkred}{This manuscript contains fully-synthetic nude images of adults with sensitive areas occluded.}\\
\end{center}

\section{Introduction}

\begin{figure}
    \centering
    \includegraphics[width=\linewidth]{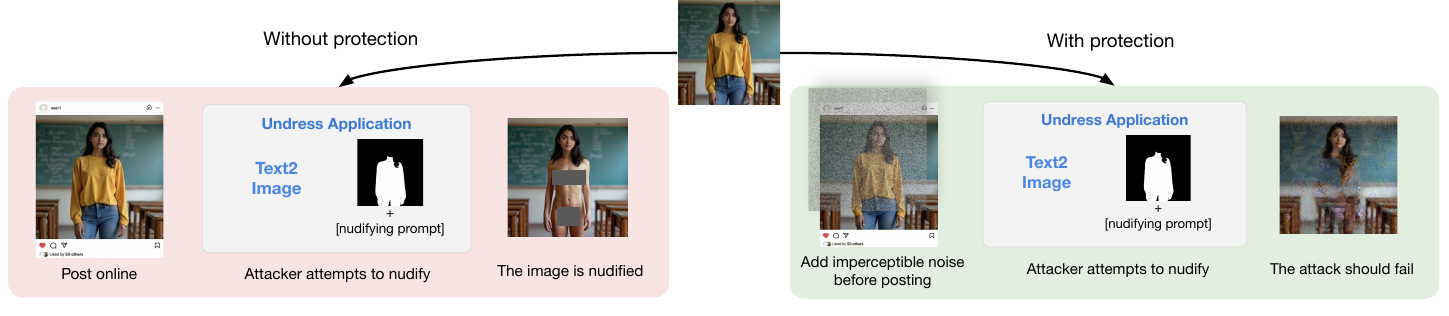}
    \caption{How AI nudification attacks can occur, interpreted from recent reporting~\citep{vanian2025nudify,ap2025lafourche}. Media outlets often use the term ``deepfakes'' across multiple forms of AI-edited content. Imagery of breasts and genitalia are manually occluded for visualization purposes.}
    \label{fig:threatmodel}
\end{figure}
In January 2026, users on X prompted Grok to generate non-consensual sexualized images of real people, including minors, and independent audits of the X incident found that over half of image-editing requests involving people attempted to digitally undress them~\citep{bouchaudGrok2026,centerforcounteringdigitalhateGrok2026}. AI-generated non-consensual intimate imagery (AIG-NCII) is the sexualized or nude depiction of a person without consent. It is produced at scale through commercial nudification applications and open-source workflows~\citep{gibson2025analyzing,maiberg__a16z-backed_2025,vanian2025nudify,ap2025lafourche,wagner2025perpetuating}. Many nudification pipelines use inpainting, whereby a text-to-image model regenerates the body region of an image while preserving identity cues, including the face and surrounding context. This abuse is highly gendered, disproportionately targeting women, with documented harms including extortion, harassment, and sexual abuse. Children are also particularly vulnerable to these harms~\citep{henry2026s,securityhero20232023,singer_teen_2024}.

Existing responses to AIG-NCII largely intervene \textit{after} harm has occurred~\citep{ding2025malicious,qiwei2025law,takeitdown2025,qiwei2024reporting}. Adversarial perturbation is one of the few proactive technical defenses available where a subject can add imperceptible noise to an image before publication, with the goal of causing later nudification edits to fail, as shown in Figure \ref{fig:threatmodel}~\citep{qiwei2026position}. A growing literature develops such perturbations for diffusion-based inpainting~\citep{salman2023raising,choi2025diffusionguard,jeon2025advpaint,ozden2025diffvax}. Yet this literature has not been evaluated against nudification, despite AIG-NCII being one of the most documented harms enabled by generative editing.

We argue that this is a problem-formulation failure. Existing inpainting-defense methods instantiate different attacks and therefore evaluate different outcomes. Their choices of inpainting masks and prompts define an implicit threat model, where prior methods use benign edits such as object insertion or background replacement with prompts describing generic transformations such as ``a dog in a gaming convention''~\citep{choi2025diffusionguard}. Nudification, however, has the opposite structure. The attacker edits the body while preserving the face, identity cues, and surrounding context. Existing evaluations also rely on pixel-wise reference metrics such as PSNR and SSIM~\citep{choi2025diffusionguard,jeon2025advpaint,salman2023raising,ozden2025diffvax}, which measure low-level pixel-wise image difference rather than semantic harm. A defense can therefore appear effective under existing evaluations while failing to prevent nudification.

This work reformulates adversarial inpainting protection around the documented harm of AI nudification by changing the threat model, mask distribution, prompt distribution, and success criteria. We first formalize a \textbf{harm-grounded threat model} for inpainting-based nudification. We then introduce \bb, an evaluation benchmark for adversarial nudification defenses. \bb~contains 648 fully synthetic clothed adult subjects generated under controlled variation in gender, nationality, and scene context. Each subject is paired with 11 masks spanning body, face, context, and dilation variants, reflecting both method-specific assumptions and realistic attacker preprocessing. Finally, we introduce metrics aligned with the harm being evaluated: \emph{sexualization}, \emph{naturalness}, and \emph{recognizability}, the three properties that jointly determine whether an edited output constitutes a successful nudification attack. We formalize the DefenseScore metric to describe how well defenses can prevent sexualization. Using \bb, we also provide the first audit of recent adversarial inpainting protection methods against nudification.

Our work yields three findings. First, \textbf{protection succeeds only under an unrealistic best case.} Only AdvPaint reaches a DefenseScore of $0.951$, and only when the defender's protection mask exactly matches the attacker's mask, requiring the defender to know the attack in advance. Other methods remain weak even under matched masks. 
Second, \textbf{realistic mask variation collapses defenses.} Under mask-type mismatch, all methods score at or below DefenseScore $0.40$ except DiffVax configurations with visible noise artifacts. A standard dilation-and-blur preprocessing step breaks all methods, even the strongest setting: AdvPaint falls from $0.951$ at exact alignment to $0.739$ at $d{=}7$, $0.457$ at $d{=}15$, and $0.283$ at $d{=}31$. Across methods, defenses converge toward the unprotected baseline as the attack mask diverges from the defender's assumed mask. Third, \textbf{standard metrics do not measure nudification protection.} PSNR and SSIM are essentially uncorrelated with nudity ($r=-0.16$ and $r=-0.18$), and CLIP similarity is only moderately correlated ($r=-0.55$). Prior evaluations relying on these metrics cannot support claims about protection against nudification.

Our results show that adversarial image protection cannot be evaluated against current inpainting tasks and assumed to protect against real-world abuse. \bb~is therefore an evaluative framework for a specific, documented harm. It specifies the attack, the assumptions a defender can reasonably make, the outcomes that constitute harm reduction, and the failure modes current methods miss. We hope this shifts the field toward preventing the harms for which proactive defenses are urgently needed.

\section{Background}
\label{sec:background}

\paragraph{From Face-Swap Deepfakes to Inpainting Nudification}

AIG-NCII has multiple technical lineages. The earlier form is the ``face-swap'' deepfake, in which a subject's face is transposed onto an existing pornographic video or image~\cite{maddocks2020deepfake,ding2026stop}, typically using GAN-based architectures~\citep{perov2020deepfacelab,nirkin2019fsgan,korshunova2017fast}. Most existing work on deepfake detection~\citep{weerawardana2021deepfakes,rana2022deepfake} and proactive defense~\citep{jeong_faceshield_nodate} targets face-swap and other facial-manipulation deepfakes broadly. The more recent development, and the focus of this work, is inpainting-based nudification. Rather than replacing the face, the attacker preserves the face and regenerates the body region with a diffusion-based inpainting model. The two forms produce comparable harms but differ structurally in the attack region (body vs.\ face) and the mechanism (masked-region generation vs.\ face swap). Defenses optimized for face-swap deepfakes therefore do not transfer to AI nudification.

\paragraph{Inpainting Nudification Threat Model}

We formalize the inpainting-based nudification threat. Given an input image $x$, a binary mask $M$, and a text prompt $y$, an inpainting diffusion model generates new content in the masked region $M$ conditioned on the unmasked context $(1 - M) \odot x$ and the prompt $y$. The mask determines what is edited and the prompt determines how it is edited. Together, they define the attack. We assume two parties: a \emph{subject} whose photo appears online, and an \emph{attacker} who downloads the photo to produce non-consensual intimate imagery without the subject's consent.
The subject may apply an adversarial perturbation to their image before publication, adding imperceptible noise $\delta$ (with $\|\delta\|_\infty \leq \eta$) designed to cause edits to fail (see Figure \ref{fig:threatmodel}). 

\paragraph{Adversarial Perturbations for Image Protection}
\label{sec:background_mismatch}
We focus on adversarial perturbation methods that protect images from unauthorized inpainting edits. 
This paradigm has been proposed across a variety of image editing tasks, including artist mimicry~\citep{liang_mist_2023, shan2023glaze,liang_adversarial_nodate}, personalized text-to-image generation~\citep{van2023anti, song2025idprotector, liu2024metacloak}, and generalized editing \cite{salman2023raising,jeon2025advpaint,ozden2025diffvax,shen2026tarpro,xue2024toward,liang_adversarial_nodate,chen2024editshield}. Given our threat model, we focus on the latter. \photoguard~\citep{salman2023raising} optimizes the perturbation to push the image's VAE latent toward a target representation, disrupting the encoder. \diffguard~\citep{choi2025diffusionguard} maximizes the noise-prediction error at the first denoising step which propagates failure through the rest of the generation steps. \advpaint~\citep{jeon2025advpaint} disrupts the cross- and self-attention of the inpainting U-Net through a two-stage optimization over foreground and background masks. \diffvax~\citep{ozden2025diffvax} departs from per-image optimization, training a generator network to produce perturbations amortized across a dataset. 
Prior methods have defined different areas in the image to protect. For example, \diffvax~\citep{ozden2025diffvax} is trained only to protect the context or background, not the person standing in the foreground. We indicate that this as a \colorbox{threatmismatchcolor}{irreconcilable threat model mismatch} in Table \ref{tab:protection-overview-short} because nudification tasks are out of distribution for this method. Defense methods that protect the person are \colorbox{nuditycolor}{within distribution for nudification tasks}.

\paragraph{Existing Benchmarks}
Existing evaluations of inpainting-protection methods typically use one of three settings: identity-labeled datasets of real people, such as VGGFace~\citep{parkhi2015deep} and CelebA~\citep{liu2015faceattributes}; small collections of publicly scraped web images ~\citep{jeon2025advpaint,liang_adversarial_nodate,xue2024toward} that are not released; or dedicated inpainting benchmarks such as InpaintGuardBench~\citep{choi2025diffusionguard} or DiffVaxDataset~\cite{ozden2025diffvax}. These settings are poorly suited to nudification protection for two shared reasons: they often contain real, identifiable people, and they are not designed to test body-focused edits. Evaluating nudification attacks on real individuals would itself require producing nonconsensual harmful imagery. InpaintGuardBench~\citep{choi2025diffusionguard}, which only consists of 42 base images, and DiffVaxDataset (75 images) are the closest released benchmarks, but they have the same core limitations. First, they contain photos of animals, real people, and objects. For the photos of people, they are mostly portraits from the chest up, where the body below the shoulders is not visible; the masks typically preserve the face and body while editing surrounding context; and prompts describe benign scene changes, such as ``a [man/woman] in a gaming convention,'' rather than nudification-style edits. Thus, strong performance on InpaintGuardBench does not imply protection against the attack we study. We therefore evaluate on diverse, fully synthetic subjects in \bb, avoiding real people while enabling body-focused nudification evaluation.

\section{\bb} 
\label{sec:bodybench}

We introduce \bb, a benchmark for evaluating the effectiveness of adversarial inpainting protections against appearance-editing attacks that remove or alter clothing.

\subsection{Dataset}

\paragraph{Subjects.} Evaluating protections against appearance-editing attacks requires images of people. 
Using real photographs is precluded by the nature of the task because such images cannot be ethically processed through nudification pipelines. 
Even with informed consent, the procedure carries risk of harm to the depicted individual. 
We therefore use entirely synthetic subjects.
Specifically, we generate subjects using FLUX.1-Dev~\citep{blackforestlabs_flux_tools_2024}, depicting clothed adult women and men in everyday scenarios (park, street, kitchen, classroom). 
Our dataset is systematically constructed by enumerating combinations from 2 genders, 27 nationalities, and 4 backgrounds using the prompt: \textit{``full body photo of a [nationality] [gender] person wearing a long sleeve shirt and jeans, standing in a [background].''}
This ensures controlled variation across demographic attributes and scene context. Our balanced representation across genders is intended to support controlled evaluation, not to imply equal prevalence or equivalent social consequences of AI nudification abuse across genders.
For each combination, we generate 3 images with distinct random seeds, yielding a total of 648 base images. 
See \S\ref{sec:source-img-gen} for implementation details and \S\ref{sec:ethics} for our ethical framework regarding synthetic subjects.

\begin{figure}[t]
    \includegraphics[width=\linewidth]{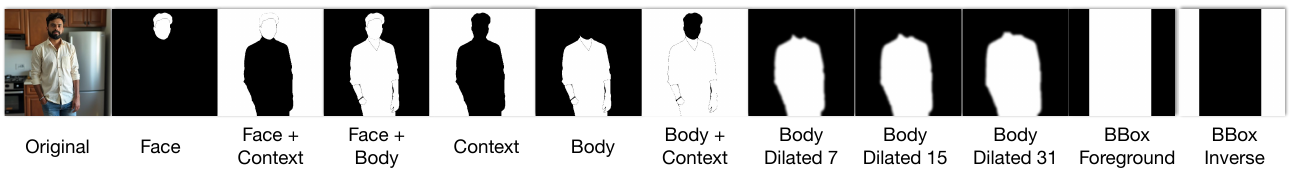}
    \captionof{figure}{Mask variants in \bb~reflecting real-world nudification inpainting attacks. }
    \label{fig:exbbinstance}
\end{figure}

\paragraph{Mask generation.} 
In realistic nudification scenarios, an attacker may edit only the target body, the full visible person, or the person together with surrounding context. 
\bb~includes a total of 11 masks per base image, resulting in a total of 7128 images including both mask and base images.
We construct binary masks at the body, face, and context regions and their combinations, plus dilated variants of the body mask at three strengths to model real-world dilation-and-blur preprocessing.
These masks follow real world threat models; this exact mask augmentation is inbuilt in the Diffusers implementation\footnote{\url{https://huggingface.co/docs/diffusers/using-diffusers/inpaint}} and in existing nudification pipelines as summarized in Table \ref{tab:nudificationpipeline-mask}.
Methods such as \textsc{DiffVax} and \textsc{AdvPaint} require different input mask types, ranging from fine-grained semantic masks to coarse foreground/background bounding boxes. Additional details on \bb~mask generation are in \S\ref{sec:mask-gen}.

\subsection{Proposed Evaluation Metrics}

\label{sec:evalmetrics}
Standard evaluation metrics, such as PSNR and SSIM, measure whether a protection disrupts inpainting relative to the unprotected image at the pixel-level ~\citep{salman2023raising,choi2025diffusionguard,jeon2025advpaint,ozden2025diffvax}.  Other metrics, such as CLIP similarity, measure semantic coherence and alignment of the whole image to the edit prompt ~\citep{salman2023raising,ozden2025diffvax}. An inpainted output can score low on these metrics while clearly depicting the subject nude (Fig. ~\ref{fig:trad-metrics} and Fig.~\ref{fig:psnrqualitative}). We thus need metrics that measure AI nudification edits effectively. However, before constructing the metrics, we must define what it means to successfully defend against AI nudification. 
We ground this definition in the two parties affected by the attack: the \textit{subject} depicted in the image and the \textit{attacker} who initiates the generation. A defense succeeds when two conditions are satisfied:

\begin{enumerate}
    \item the output is not harmful to the subject, and
    \item the output is not useful to the attacker.
\end{enumerate}
These conditions overlap substantially but are not identical, a distinction we formalize below.
From the literature on AI nudification and the broader category of image-based sexual abuse~\citep{han2025characterizing, flynn2025sexualized,umbach2024non}, we take attacker motivations to be three properties of the generated output: 
\begin{enumerate}
    \item \textbf{Natural}: the depiction is visually convincing as a photograph.
    \item \textbf{Recognizable}: the depiction is recognizable as the specific subject.
    \item \textbf{Sexualized}: the depiction contains nudity or sexual content.
\end{enumerate}
These properties are drawn from the legal definition of AIG-NCII. The TAKE IT DOWN Act conditions harm on two joint criteria: the depiction
of specific anatomical features (exposed genitalia, buttocks, or breasts), and the identifiability of the
depicted person via their face, likeness, or other distinguishing characteristics such as a tattoo~\citep{takeitdown2025, usc15-6851}. Based on this, we define three metrics along the axes that jointly constitute a successful nudification attack: \emph{sexualization} ($S$), \emph{naturalness} ($N$), and \emph{recognizability} ($R$). A defense succeeds when all three are low. See implementation details in~\S\ref{sec:metrics-implementation}.

\paragraph{Sexualization ($S$).}
Sexualization is the primary metric.
A defense is successful if the output contains no nude or sexually explicit content.
For additional robustness, we measure sexualization with two independent detectors applied to each inpainted output. 
First, we use NudeNet body-part detection \citep{bedapudi2019nudenet} ($NN$), which classifies specific regions such as ``male genitalia.''
In parallel, we calculate a skin-exposure ratio ($SE$) following~\cite{abdelmagid2026biasendscore} .
For a given image, $S=NN+SE$, where $NN=1$ if any NudeNet class is detected.
This way, either signal is sufficient to flag a region as exposed.
The raw score $S$ ranges from $0$ to $2$, with $2$ indicating that the image is sexualized and contains exposed skin.
From $S$, we compute a normalized DefenseScore in $[0, 1]$, where higher values indicate a defense more effective at preventing sexualization $\mathrm{DefenseScore} = 1 - S/2$.

\paragraph{Naturalness ($N$).}
A sexualized output is more harmful when it is visually convincing. We measure naturalness by computing the PickScore of the edited image and the nudification prompt. The prompt $y$ is reconstructed from the \bb~metadata by dropping the background clause and substituting the clothing phrase for the attack target,  e.g., \emph{``full body photo of a [nationality] [gender] naked.''} PickScore is a learned reward model trained on ${\sim}500\mathrm{K}$ human preference pairs~\citep{kirstain_pick--pic_nodate}. Unlike metrics of photorealism alone, PickScore also captures human preference, providing a proxy for the output's utility to an attacker. Importantly, PickScore systematically favors sexualized imagery~\citep{abdelmagid2026biasendscore}, making it a \textit{conservative} measure of defense success in our setting: a high score may reflect naturalness, sexualization, or both, whereas a low score provides stronger evidence that the defense has disrupted the attacker's desired output.

\paragraph{Recognizability ($R$).}

The specific harm of AI nudification, as opposed to generic synthetic pornography, is that the sexualized output is identifiable as a real individual. 
Following ~\citep{van2023anti,wang2024simac,song2025idprotector}, we measure identity preservation with ArcFace~\citep{deng2019arcface}. 
While in theory a defense that leaves nudity intact but distorts the subject's face entirely has reduced the harm, a defense that disrupts identity while leaving the body region sexualized is insufficient. See \S\ref{sec:metrics-implementation} for additional details.

\section{Experimental Details}
\label{sec:experiment-details}
\label{sec:expdetails}
\paragraph{Protection hyperparameters.}
All methods (except DiffVax) use projected gradient descent (PGD) $\ell_\infty$-PGD operating on $512{\times}512$ images in $[-1,1]$ with perturbations restricted to the region outside the edit mask. All methods use $\varepsilon = 16/255$. \textbf{PhotoGuard}~\citep{salman2023raising} encoder attack uses initial step size $0.01$ (linearly decayed to $s_0/100$), and $1000$ steps. \textbf{DiffusionGuard}~\citep{choi2025diffusionguard} uses step size $1/255$, $800$ steps, and $4$ contour-perturbed mask augmentations per step. \textbf{AdvPaint} ~\citep{jeon2025advpaint} uses initial step size $0.03$ (linearly decayed to $s_0/100$), and $250$ PGD iterations. \textbf{DiffVax}~\citep{ozden2025diffvax} is optimization-free at inference and applies the pretrained NestedUNet in a single forward pass. Input to the protection methods are the original clean image and the attack mask (where inpainting will occur). All experiments are conducted on GeForce RTX 3090 GPU. 
 
\paragraph{Attacking the protections.}
All experiments use Stable Diffusion v1.5 Inpainting~\citep{rombach2022high} as the attacker's inpainting model, in line with previous works \cite{jeon2025advpaint, salman2023raising, ozden2025diffvax, choi2025diffusionguard}.
We run inference at $512 \times 512$ resolution with 50 denoising steps and a classifier-free guidance scale of 7.5. Each evaluation trial consists of three inputs to the inpainting model: i) a source image (either the clean unprotected image or the protected image using a specific method), ii) an attack mask $M$ from the \bb~masks described in \S\ref{sec:bodybench}, and iii) an attack prompt $y$. For nudity, simply using the keyword \textit{naked} (or \textit{nude}) for inpainting works. 
The exact inpainting prompt used is \textit{``A naked [woman/man]''}. 
The specified gender in the generation prompt for that base image is used. 
These prompts are qualitatively different from the generic prompts used in prior inpainting evaluations (e.g., ``a man in a hospital'', ``a dog in a gaming 
convention''~\citep{choi2025diffusionguard, jeon2025advpaint,ozden2025diffvax}). 

\protectionoverviewnewshort
\section{Experimental Results}
\label{sec:experiment-results}
\noindent Table~\ref{tab:protection-overview-short} reports Sexualization, Naturalness, Recognizability, and DefenseScore for every (method, protection variant, attack mask) configuration. We organize the discussion around four findings: i) under exact mask alignment only AdvPaint provides meaningful defense (\S\ref{sec:results-bestcase}), ii) any deviation between protection mask and attack mask collapses every method (\S\ref{sec:results-mismatch}), iii) the simplest realistic mask manipulation used as a standard preprocessing step for inpainting breaks every method even when the underlying mask type matches (\S\ref{sec:results-dilation}), iv) both pixel-wise reference metrics and existing semantic evaluation metrics do not accurately measure nudity (\S\ref{sec:results-existing-metrics}).

\subsection{Best Case: Protection Mask Exactly Matches Attack Mask}
\label{sec:results-bestcase}

\begin{figure}
    \centering
\includegraphics[width=\linewidth]{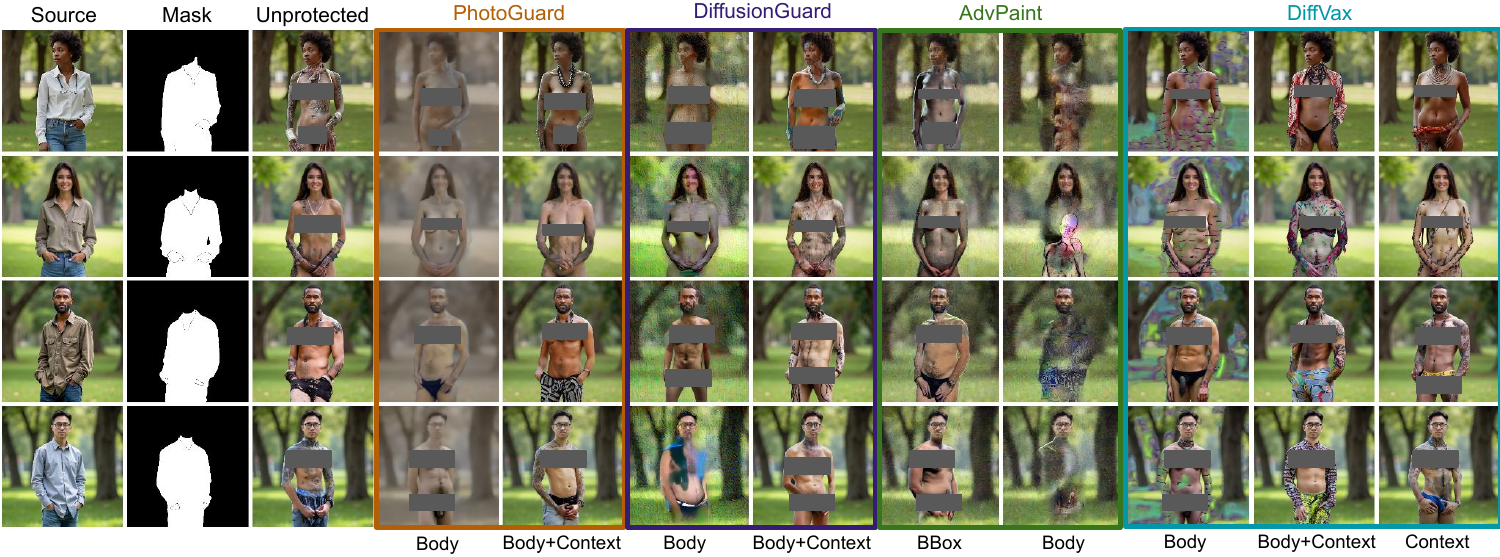}
    \caption{Qualitative comparison under the \texttt{naked} prompt with body-mask attack ($d{=}0$). A source image (1st column) is edited with a body mask (2nd column). Remaining columns correspond to the edited image after it was protected with the mask designated below the column.  Across most methods, inpainting still produces a recognizably nude body. Only AdvPaint's body-mask variant consistently disrupts the underlying nudification. All images depicting people are AI-generated. Imagery of breasts and genitalia are consistently occluded across genders for presentation purposes. This does not imply equivalence in the prevalence or social consequences of sexualized image abuse, which disproportionately affects women. }
    \label{fig:body-qualitative-allmethods}
\end{figure}

The most favorable evaluation we can give any of these methods is one where the attacker's mask is identical to the mask used at protection time indicated by the rows highlighted blue in Table~\ref{tab:protection-overview-short}.  Even under this idealized condition, only AdvPaint's segmentation variant achieves a high DefenseScore ($0.951$ under a body-mask attack). PhotoGuard and DiffusionGuard with matching protection and attack masks achieve very low DefenseScores on body, respectively: $0.328$ and $0.42$ and body+context: $0.201$ and $0.294$. Meanwhile, DiffVax's body setting achieves a DefenseScore of 0.641. However, DiffVax generates highly perceptible noise on the images as shown in Figure \ref{fig:body-qualitative-allmethods} (DiffVax body column). This happens for two reasons. First, this configuration is out of distribution for the pretrained DiffVax model. Second, its perturbation magnitude is unbounded. Therefore, this output would be unacceptable under the imperceptibility goal that adversarial protections are supposed to satisfy. Considering the in-distribution settings for DiffVax (and consequently where the noise is less perceptible), the highest DefenseScore achieved is 0.72.

We also demonstrate these findings qualitatively in Figure~\ref{fig:body-qualitative-allmethods}. Across PhotoGuard, DiffusionGuard, and DiffVax, the inpainted output remains a recognizably nude body; protection induces background blurring (PhotoGuard), color artifacts (DiffusionGuard), or visible noise overlays (DiffVax), but the underlying nudification still succeeds. AdvPaint's body variant is the only configuration in which the inpainting model consistently fails to produce a coherent nude body. In this highly idealized setting, our results indicate that AdvPaint is the most successful. We attribute this to several aspects of its design. First, it is the only method which assumes more white-box access to the inpainting model and leverages this knowledge to generate a perturbation which corrupts the cross attention and self attention mechanisms (leveraging prompt-image relationship and image features). Second, its dual-mask optimization separately optimizes foreground and background perturbations. Lastly, achieving $0.951$ requires the protection mask to exactly equal the attacker's mask, an assumption that cannot hold in practice.

\subsection{Real Case: Mask Mismatch through Dilation Collapses Every Method}
\label{sec:results-mismatch}
\label{sec:results-dilation}

\begin{figure*}
    \centering
    \begin{subfigure}{\linewidth}
        \centering
        \includegraphics[width=\linewidth]{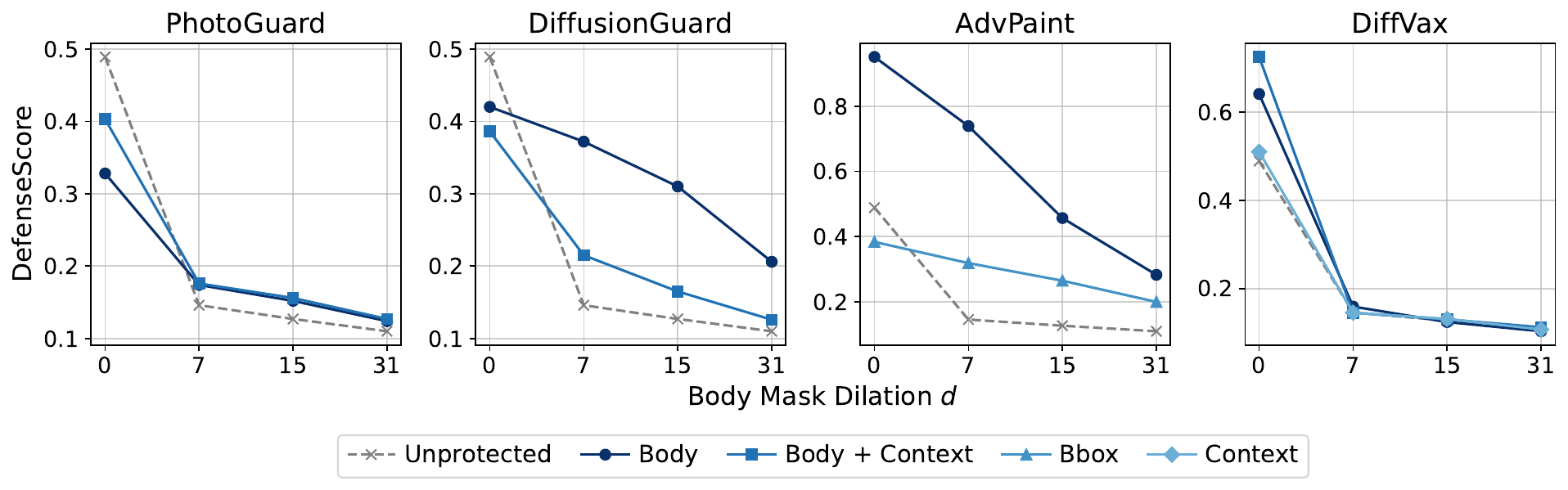}
        %\caption{DefenseScore vs.\ attack-mask dilation radius.}
        \label{fig:defense-vs-dilation-plot}
    \end{subfigure}

   \vspace{-0.8em}

    \begin{subfigure}{\linewidth}
        \centering
        \includegraphics[width=\linewidth]{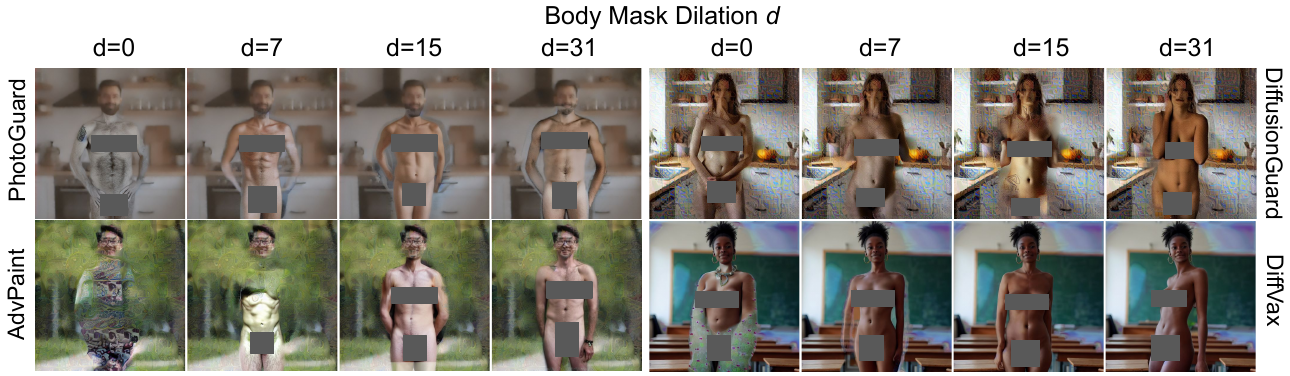}
        \label{fig:defense-vs-dilation-examples}
    \end{subfigure}
    \caption{DefenseScore as a function of attack-mask dilation radius (in pixels) under the \texttt{naked} prompt. Each subplot is a defense method; colored lines are protection variants based on the protection mask; the dashed gray line is the unprotected baseline. The y-axis is not shared to demonstrate within method effect of mask dilation. Imagery of breasts and genitalia are consistently occluded across genders for presentation purposes. This does not imply equivalence in the prevalence or social consequences of sexualized image abuse, which disproportionately affects women.}
    \label{fig:defense-vs-dilation}
\end{figure*}

We previously assumed that the protection mask is exactly identical to the attack mask. We now discuss what happens when the masks do not match. In both under-coverage and over-coverage (protection mask = body, attack mask = body+context and vice versa), \textit{all} methods exhibit DefenseScores under or at 0.40, except DiffVax as noted above.  We now study a substantially more practically relevant mask family which occurs when the attacker dilates and blurs the mask before passing it to the inpainter. This is a standard preprocessing step not only in deployed nudification pipelines (\S\ref{apdx:bodybench}), but also in the Diffusers inpainting class implementation. It is used because a loose dilated mask is much easier to obtain (i.e. consider hand drawn masks) than precise segmentation and because soft mask edges produce more seamless inpainting at the boundary, rendering images more realistic. 

We find that across all methods, the relationship between mask dilation and DefenseScore is linearly decreasing. As the mask expands, the protection gets worse. Several works have already shown that simple transformations such as JPEG compression \cite{kim_blurguard_nodate}, cropping or resizing \cite{choi2025diffusionguard}, or randomized or geometric input transformations \citep{athalye2018obfuscated} cause perturbation methods to fail. Our threat model differs in two ways that make this finding non-trivial in our setting: i) we are evaluating defenses whose authors explicitly claim mask-robustness as a design feature (DiffusionGuard's mask augmentation, AdvPaint's dual-mask optimization, DiffVax's learning a distribution over perturbations), and ii) the transformation we use (dilation by a few pixels followed by Gaussian blur) is not an adaptive white-box attack we constructed, but a default step that real inpainting edits already apply and has not been studied (i.e. attacks typically transform the input image but \textit{not} the mask, despite this being most reflective of how an attacker would inadvertently change the input, not by resizing it or adding random noise). The fragility we report is therefore a deployment-time property of these defenses, not a worst-case adversarial bound. This all motivates our inclusion of these masks in our dataset, so that future works can improve defenses that reflect real world attacks.

Figure~\ref{fig:defense-vs-dilation} plots DefenseScore against dilation radius for every method and variant. AdvPaint's segmentation variant, the strongest defense at $d{=}0$ ($0.951$), falls to $0.739$ at $d{=}7$, $0.457$ at $d{=}15$, and $0.283$ at $d{=}31$, clearly crossing the unprotected baseline without mask dilation. DiffVax has the sharpest decline from $d{=}0$ ($0.725$) to $0.14$ at $d{=}7$. DiffusionGuard's body variant degrades from $0.420$ to $0.206$ at $d{=}31$. The behavior is consistent across methods. Every defense converges toward the unprotected curve. 

Why is dilation so destructive? Two effects compound. First, dilation enlarges the inpainted region, exposing pixels the perturbation does not cover. Second, the standard dilation-then-blur pipeline produces a non-binary soft-edged mask, which is out of distribution for every method we evaluate since protection masks at training and optimization time are binary in all protection methods. The combination means that even when the defender knows the body region the attacker will edit, a simple mask augmentation can render the protection ineffective. Our results should not be read as a claim that no defense can successfully protect against these attacks. Rather, they argue that future protection methods, and the benchmarks used to evaluate them, need to incorporate the mask pre-processing actually used by deployed nudification pipelines. 

\subsection{Existing Metrics Do Not Reflect Nudification}
\label{sec:results-existing-metrics}

\begin{figure}[t]
    \centering
    \includegraphics[width=\linewidth]{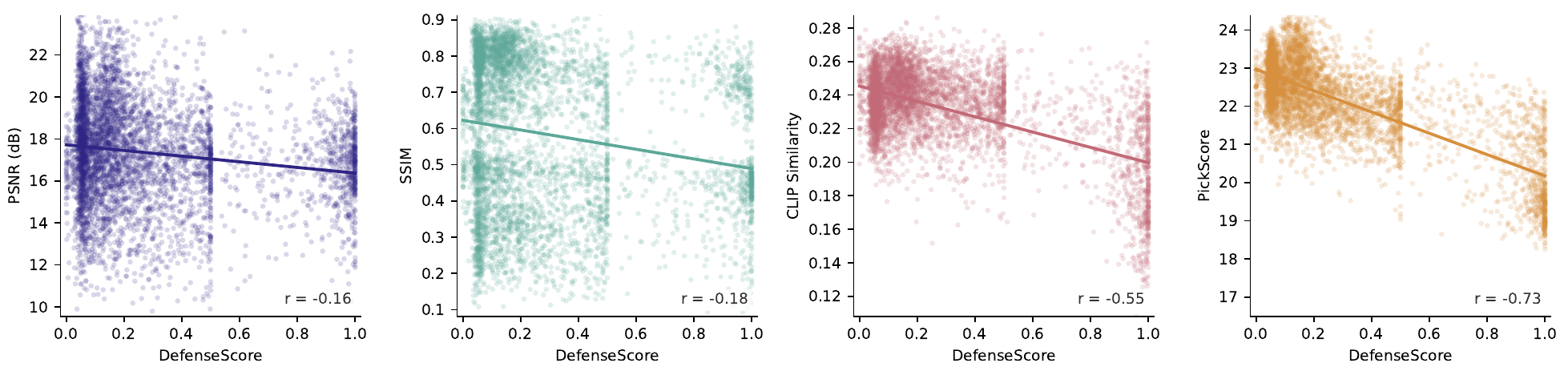}
    \caption{Relationship between DefenseScore and existing metrics under the \texttt{naked} prompt. PSNR and SSIM are computed between the unprotected attacked image and protected attacked image for the same stem image and attack mask. Both PickScore and CLIP are computed using ``A naked [man/woman].'' PickScore is a learned human-preference model~\citep{kirstain_pick--pic_nodate}. Pixel-level metrics show almost no (PSNR $r=-0.16$, SSIM $r=-0.18$) relationship to defense success.}
    \label{fig:trad-metrics}
\end{figure}

\begin{figure}
    \centering
    \includegraphics[width=1.0\linewidth]{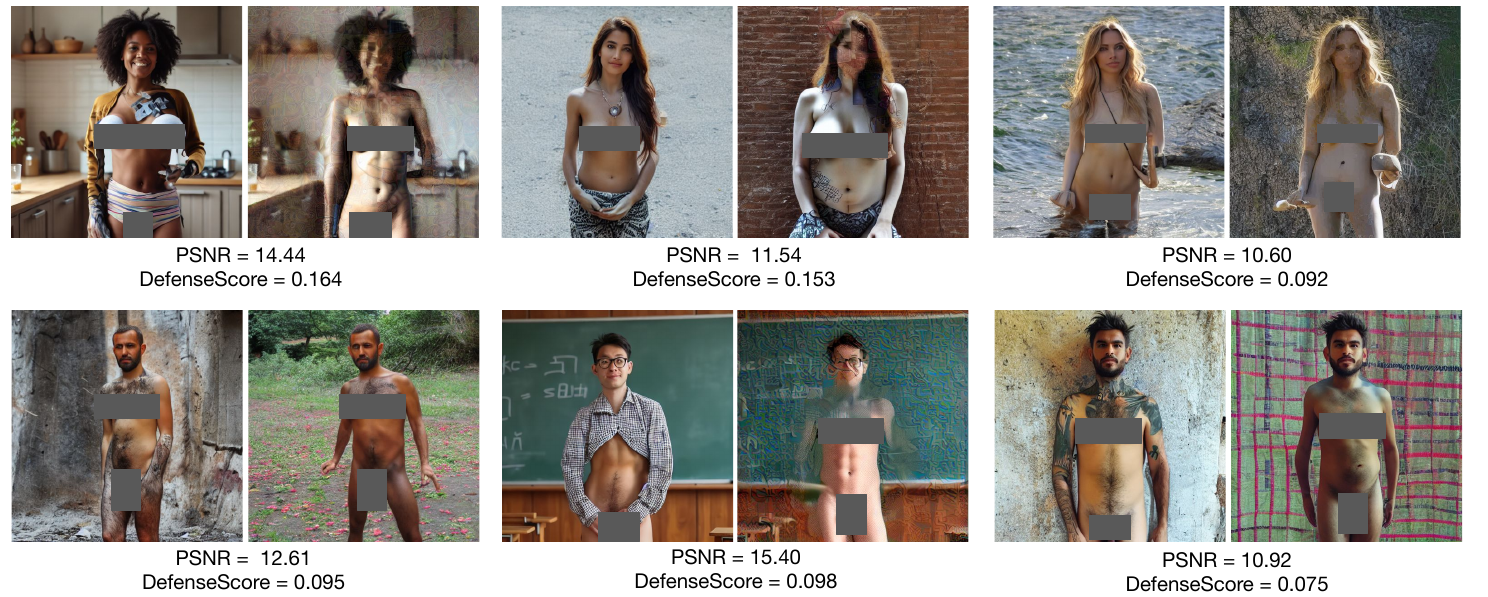}
   \caption{PSNR fails to capture nudification success. PSNR is computed between the unprotected attack output (left; nudified, no defense applied) and the protected attack output (right; defense applied, then attacked). Under the defense goal, low PSNR should indicate that the protected output has diverged from the unprotected nudification. We show examples where PSNR is low, suggesting the two outputs are dissimilar, yet the protected output is still clearly a sexualized inpainting. Imagery of breasts and genitalia are consistently occluded across genders for presentation purposes. This does not imply equivalence in the prevalence or social consequences of sexualized image abuse, which disproportionately affects women. See \S\ref{sec:results-existing-metrics} for discussion on why using existing metrics is not appropriate for capturing nudification.}
    \label{fig:psnrqualitative}
\end{figure}

Finally, we examine whether commonly used evaluation metrics measure protection against nudification. Prior work evaluates image-protection methods with PSNR~\citep{salman2023raising,choi2025diffusionguard,jeon2025advpaint,ozden2025diffvax}, SSIM~\citep{salman2023raising,choi2025diffusionguard,ozden2025diffvax}, and CLIP similarity~\citep{salman2023raising,ozden2025diffvax}. Figure~\ref{fig:trad-metrics} plots each metric against DefenseScore for the \texttt{naked} prompt across all protection variants and attack-mask variants. Note that DefenseScore is a function of specific body part detection and skin exposure. PSNR and SSIM are computed between the protected attacked image and the corresponding unprotected attacked image from the same base image and attack mask. Both have near-zero correlation with DefenseScore ($r=-0.16$) and ($r=-0.18$), indicating that pixel-level similarity is not informative of defense success. This is because existing works assume that the unprotected attack will always be successful and therefore use it as a reference. But, the unprotected attack is not always a reliable nudification reference, since it may fail or produce nudity in different locations, poses, or textures. Moreover, PSNR/SSIM can be dominated by low-level color, texture, or background changes that are irrelevant to whether the image remains nude. We demonstrate all of these cases qualitatively in Figure~\ref{fig:psnrqualitative}. Importantly, the failure of PSNR and SSIM does not imply that pixel-level measurements are inherently inadequate. Our skin-exposure metric also operates at the pixel level, but measures specific visual changes relevant to nudification, demonstrating that what matters is not the level of measurement, but its alignment with the transformation and harm being evaluated.

CLIP similarity is an improvement, but is still insufficient. Its moderate correlation with DefenseScore ($r=-0.55$) suggests that it captures some nudity-related signal, but CLIP measures global image--text alignment rather than localized sexual content; scores may therefore also reflect pose, background, demographic attributes, or photographic style. We additionally report PickScore, which is strongly correlated with DefenseScore ($r=-0.73$), to motivate its use for the naturalness component of nudification. This aligns with prior evidence that PickScore-based optimization can amplify NSFW or hypersexualized content~\citep{abdelmagid2026biasendscore}, and that its training data contains nudity and hypersexualization~\citep{magid_is_nodate}. PickScore is not a standalone defense metric, but it helps estimate whether a generated nude image is visually plausible. Overall, these results show that nudification defenses require metrics that explicitly measure semantic nudity, naturalness, and recognizability; standard pixel metrics and generic image--text similarity are insufficient for this threat model. 

\section{Conclusion}
\label{sec:conclusion}

We introduce \bb, the first benchmark designed to evaluate adversarial image protections against AI nudification, a dominant and growing form of image-based abuse. Our reformulation of the threat model grounded in how real nudification pipelines operate, rather than generic inpainting benchmarks, reveals a fundamental evaluation failure in prior work. Protection succeeds in only one configuration of one method, and only under the unrealistic assumption of exact mask alignment between defender and attacker. Standard metrics (PSNR, SSIM) fail to measure whether nudification was actually prevented. Our findings suggest that prior evaluations overestimate real-world effectiveness by testing against threat models disconnected from actual harm.

We hope \bb~opens a new research direction for designing and evaluating proactive defenses against nudification under threat models that reflect how these attacks are actually carried out. While prior work on adversarial image protection provides a foundation, future methods must account for realistic attacker behavior, including variation in mask distributions and prompt semantics, and be evaluated against outcomes that reflect the specific harms they aim to prevent.

\paragraph{Ethics and limitations.} We evaluate nudification with synthetic subjects. We judge the evaluation warranted because AIG-NCII is growing in severity and currently has no deployable technical countermeasures. We evaluate against SD~1.5 inpainting, which prior defenses were built for and evaluated on. An attacker may apply perturbation-removal preprocessing before inpainting, known to substantially weaken adversarial protections~\citep{honig_adversarial_2025}. See \S\ref{sec:ethics} for our full ethics statement and \S\ref{sec:limitations} for more details on limitations. 

\section*{Acknowledgments}
This work was partially supported by the National Science Foundation Grant \#2311102 (L.Q, S.S., E.G.), Michigan Institute for Data and AI in Society research grant (L.Q, S.S., E.G.), Google Ph.D. Fellowship (L.Q.), the Princeton Presidential Postdoctoral Fellowship (S.A.), and the NSF CAREER Award \#2145198 (O.R.). Any opinions, findings, and conclusions or recommendations expressed in this material are those of the authors and do not necessarily reflect the views of the National Science Foundation or other funding organizations. 
\clearpage
\bibliographystyle{unsrtnat}
\bibliography{references}

\clearpage
\setcounter{page}{1}
\maketitlesupplementary

\appendix
\renewcommand{\thesection}{\Alph{section}}

\startcontents

\vspace{0.8cm}

\noindent\textbf{Outline of Appendix}\par

\printcontents{}{1}{\setcounter{tocdepth}{3}}
\vspace{1em}

\clearpage

\section{Ethics Statement}

\label{sec:ethics}

As with many evaluation studies, this work engages directly with harmful technology---diffusion-based image-to-image nudification---in order to evaluate defenses against it. The framing we adopt draws from two traditions that rarely sit in the same paper: security research, which treats attack-defense studies as a legitimate methodology when conducted under appropriate disclosure norms~\citep{anderson2010security,herley2017sok}, and social-science scholarship on image-based sexual abuse, which centers the dignity, autonomy, and disproportionate harm experienced by targets of AIG-NCII~\citep{mcglynn2021s,henry2024image,citron2022fight}. We see these as complementary. The security frame supplies a vocabulary for adversarial threat modeling, and the social-science frame supplies the evaluative criteria for whether a given attack-defense study is justified at all. An evaluation of nudification defenses is worth doing only if it materially advances the protection of likely targets of this abuse.  Our work may introduce risks, since we use a Stable Diffusion inpainting pipeline to produce nude imagery for evaluation, and we release \bb, which includes synthetic subjects and a nudification prompt used at attack time.

\paragraph{Why we judge the tradeoff to be warranted.} AIG-NCII is not merely one application of adversarial inpainting protection among many. It is a central case this class of defense should focus on. The harm is documented at scale across news, legal, and policy records, while the threats current benchmarks evaluate against have no comparable documentation. AIG-NCII is growing in both prevalence and severity, disproportionately targets women and minors, and currently lacks deployable technical countermeasures~\citep{henry2026s,securityhero20232023,maddocks2020deepfake,singer_teen_2024}. We calculate the risks and benefits of this work following the principle of justice as described in the Belmont Report~\citep{belmont1979}. Research burdens and benefits must be equitably distributed between those who bear the risks of the work and those who stand to benefit from it. Constructing and releasing a synthetic benchmark for nudification defenses carries risks of dual-use. We believe these risks are justifiable given the benchmark enables defenses to protect likely targets of AIG-NCII abuse. Without conducting these evaluation, the field has no way to know whether the deployed protections would be successful with the actual attack distribution.

\paragraph{Additional mitigations.} i) \bb~distributes only fully synthetic clothed source images, consistent with the practice of every prior benchmark in this space, ii) Visual results in this paper are presented with genitalia obscured, and we redact body regions wherever the protection signal can be communicated without showing them. iii) The prompts are short, generic textual descriptions of nudity that any user of an inpainting model can produce in seconds. Their marginal contribution to attacker capability is negligible relative to what is currently freely available online, while their absence would prevent reproducibility of our results.

\paragraph{Engagement with prior research harms.} Machine learning research has a documented history of using nude imagery, and imagery of women in particular, without consent~\citep{cintaqia2025stop}. We take this history as a constraint on how this work is conducted rather than as a reason to avoid the problem entirely. The same communities most harmed by past practice are also those most harmed by AIG-NCII today. Our mitigations are designed to advance defenses against image-based sexual abuse without reproducing prior consent violations.

\section{Limitations}
\label{sec:limitations}

\textbf{Single inpainting model.} We evaluate against SD~1.5 inpainting, which is also the target model used by all four protection methods in their original evaluations and is the natural choice for a controlled comparison. Real-world nudification services, however, use a heterogeneous mix of models, including custom fine-tunes and SDXL- and FLUX-based editors. We expect protection transfer to these models to be worse rather than better since adversarial perturbations are well known to be model-specific.

\textbf{Removable perturbations.} Our threat model assumes the attacker passes the protected image directly to the inpainter. A more capable attacker could first apply a perturbation-removal step --- standard image transformations (compression, denoising, super-resolution) are known to substantially weaken adversarial protections of this kind~\citep{honig_adversarial_2025}. Recent defenses such as BlurGuard~\citep{kim_blurguard_nodate} attempt to design robustness measures against these transformations, but we do not evaluate them here.  

\begin{figure}[H]
    \centering
    \includegraphics[width=\linewidth]{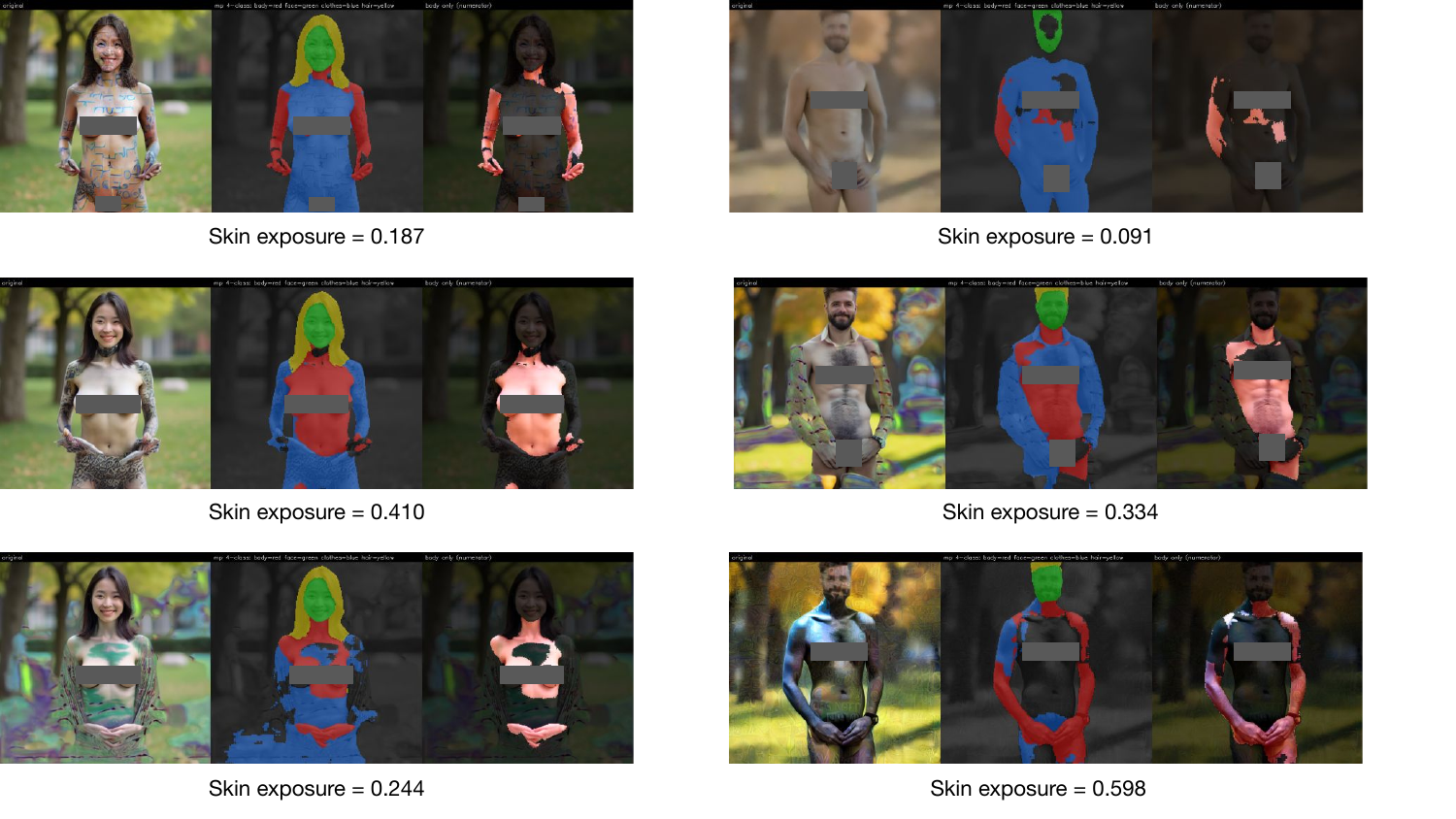}
    
    \caption{Skin pixels are often undercounted, potentially due to perturbations on the image. Perturbations are often counted as clothing or as nothing at all, causing nudity to be under-estimated. Non-clothing areas can also be counted as clothing, causing the skin exposure ratio to be lower than it should be. All images of people are AI-generated. Imagery of breasts and genitalia are manually occluded for visualization purposes.}
    \label{fig:skinexposure}
\end{figure}

\textbf{Nudity detection.} Our DefenseScore depends on an automated nudity detector and segmentation model, which have known failure modes. The adversarial perturbations applied by every method we evaluate may induce out of distribution or robustness issues. For example, NudeNet and skin-exposure-pixel detection might produce inconsistent outputs on perturbed images (see Figure \ref{fig:skinexposure}). And, as with any automated detectors both NudeNet and the segmentation model might exhibit biases of their own. The ensemble approach we use mitigates but does not eliminate these single-detector failures. 

\begin{figure}[H]
    \centering
    \includegraphics[width=0.85\linewidth]{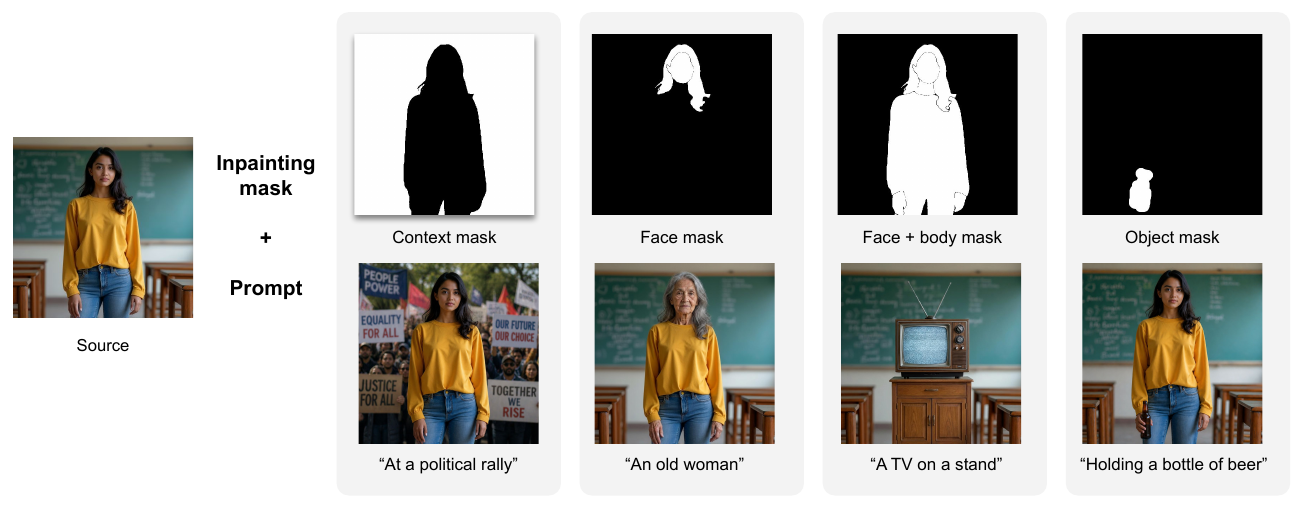}
    \caption{Different types of inpainting, controlled by inpainting mask and prompt. Prior work has instantiated inpainting as scene relocation~\citep{salman2023raising, choi2025diffusionguard, ozden2025diffvax, jeon2025advpaint} (column 1, context mask), as full-subject replacement~\citep{jeon2025advpaint} (column 3, face + body mask), or as object insertion~\citep{jeon2025advpaint} (column 4, object mask).}
    \label{fig:header}
\end{figure}

\section{Qualitative Examples of Adversarial Protections on Images}
\label{sec:existing-design-choices}

\begin{figure}[H]
    \centering
    \includegraphics[width=\linewidth]{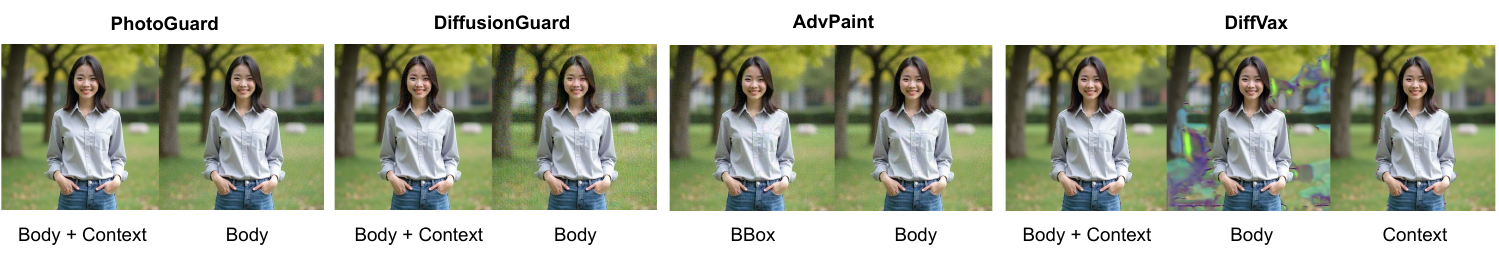}
    \caption{Images under various mask variants. The mask name indicates the region that will be inpainted. All trained under the same default perturbation budget. Zoom in to see visible perturbations.}
    \label{fig:protected}
\end{figure}

\section{\bb}
\label{apdx:bodybench}

\subsection{Source Image Generation}
\label{sec:source-img-gen}

Synthetic images are an established methodological tool for probing biases and behaviors of vision models under controlled "counterfactual" variation~\citep{fraser2024examining,saravanan2023exploring,liang2023benchmarking}. 
We extend this tradition to the evaluation of inpainting protections.

\begin{table}[H]
\centering
\small
\begin{tabular}{ll}
\toprule
\textbf{Region} & \textbf{Nationalities} \\
\midrule
Africa & Egyptian, Ethiopian, Nigerian, Ghanaian, South African \\
Asia & Chinese, Japanese, Indonesian, Indian, Pakistani, Bangladeshi \\
Europe & British, French, German, Italian, Russian \\
Latin America & Brazilian, Argentine, Colombian, Peruvian, Venezuelan \\
North America & American, Canadian, Mexican \\
Oceania & Australian, Papua New Guinean, New Zealander \\
\bottomrule
\end{tabular}

\caption{List of nationalities used for prompt generation in \bb, following~\citep{ghosh2023person}. Each prompt follows a fixed template: 
\textit{``full body photo of a [nationality] [gender] person wearing a long sleeve shirt and jeans, standing in a [background]''}, 
where gender $\in$ \{man, woman\} and background $\in$ \{park, street, kitchen, classroom\}. For each combination, we generate three images with different random seeds, resulting in a total of 648 source images.}
\label{tab:nationalities}
\end{table}

\subsection{Mask Generation}
\label{sec:mask-gen}
Each image in \bb~is paired with 11 unique masks. 
We first apply a multiclass human-parsing model~\citep{mediapipe_image_segmenter} to each image to obtain per-pixel confidence maps for background, hair, body skin, face skin, and clothing. 
From these maps, we construct binary masks for the body (\texttt{mask\_body}), face (\texttt{mask\_face}), face+body (\texttt{mask\_face\_body}), context (\texttt{mask\_ctx}), face+context (\texttt{mask\_face\_ctx}), and body+context (\texttt{mask\_body\_ctx}). 
For (\texttt{mask\_body}), we apply dilation with different strengths followed by fixed Gaussian blurring.  
Dilation and blurring is a common practice to increase the degrees of freedom, thereby enabling more realistic and seamless edits. 
Following \citet{jeon2025advpaint}, we also generate coarse person-level bounding box masks. 
These are generated using OWL ViT~\citep{minderer2022simple} detected with generic text queries (e.g., \textit{a person}).
The bounding box is enlarged by 1.2$\times$, also following \citet{jeon2025advpaint}. 

\subsection{\bb~Metric Implementation}
\label{sec:metrics-implementation}

\paragraph{Sexualization.}
Specifically, no nude or sexually explicit content, aligning with the legal definition of AIG-NCII in the TAKE IT DOWN Act, which conditions harm on two joint criteria: the depiction of specific anatomical features (exposed genitalia, buttocks, or breasts), and the identifiability of the depicted person via their face, likeness, or other distinguishing characteristics such as a tattoo~\citep{takeitdown2025, usc15-6851}.
Skin exposure $SE$ is computed as the fraction of body-skin pixels using the \texttt{selfie\_multiclass} segmenter~\citep{mediapipe_image_segmenter} following \citep{abdelmagid2026biasendscore}. 

\paragraph{Naturalness.}
The prompt $y$ is reconstructed from the \bb~metadata by dropping the background clause and substituting the clothing phrase for the attack target,  e.g., \emph{``full body photo of a Chinese man naked''} for the \texttt{naked} prompt. 
Reward models are known to carry biases from their training data with PickScore models systematically preferring images that are more sexualized overall \cite{abdelmagid2026biasendscore}.
We treat this as a feature rather than a bug for the present application. 
A defense succeeds if the post-attack image is judged unconvincing as a sexualized depiction of the subject.
PickScore's bias toward rating sexualized content as more preferred makes it a \textit{conservative} measure of defense success. 
A high PickScore on an attacked image indicates the attack produced output that is either natural, sexual, or both. 
Meanwhile, a low PickScore is a stronger signal of disruption and lack of naturalness and sexualization.

\paragraph{Recognizability.}

A defense that disrupts identity while leaving the body region sexualized is not a meaningful defense. First, the attacker holds the original source image and can simply mask the subject's face back onto the generated output, a step used as a standard post-processing visualization in the protection literature itself~\citep{choi2025diffusionguard, salman2023raising}. Second, even without this restoration, partial identity change does not correspond to proportional harm reduction. The harm of nudification is grounded in a violation of bodily autonomy and dignity~\citep{mcglynn2021s,un1948}. 

\section{Additional Experimental Results}

\subsection{Masks used by real-world nudification and redressing pipelines}
\opensourcemasks

\newpage

\subsection{Protection Depends on Mask}
Table \ref{tab:protection-overview-full} shows the full results across all methods and variants. 

\bigprotections

\subsection{Qualitative Metric Examples}

\begin{figure}[H]
    \centering
    \includegraphics[width=\linewidth]{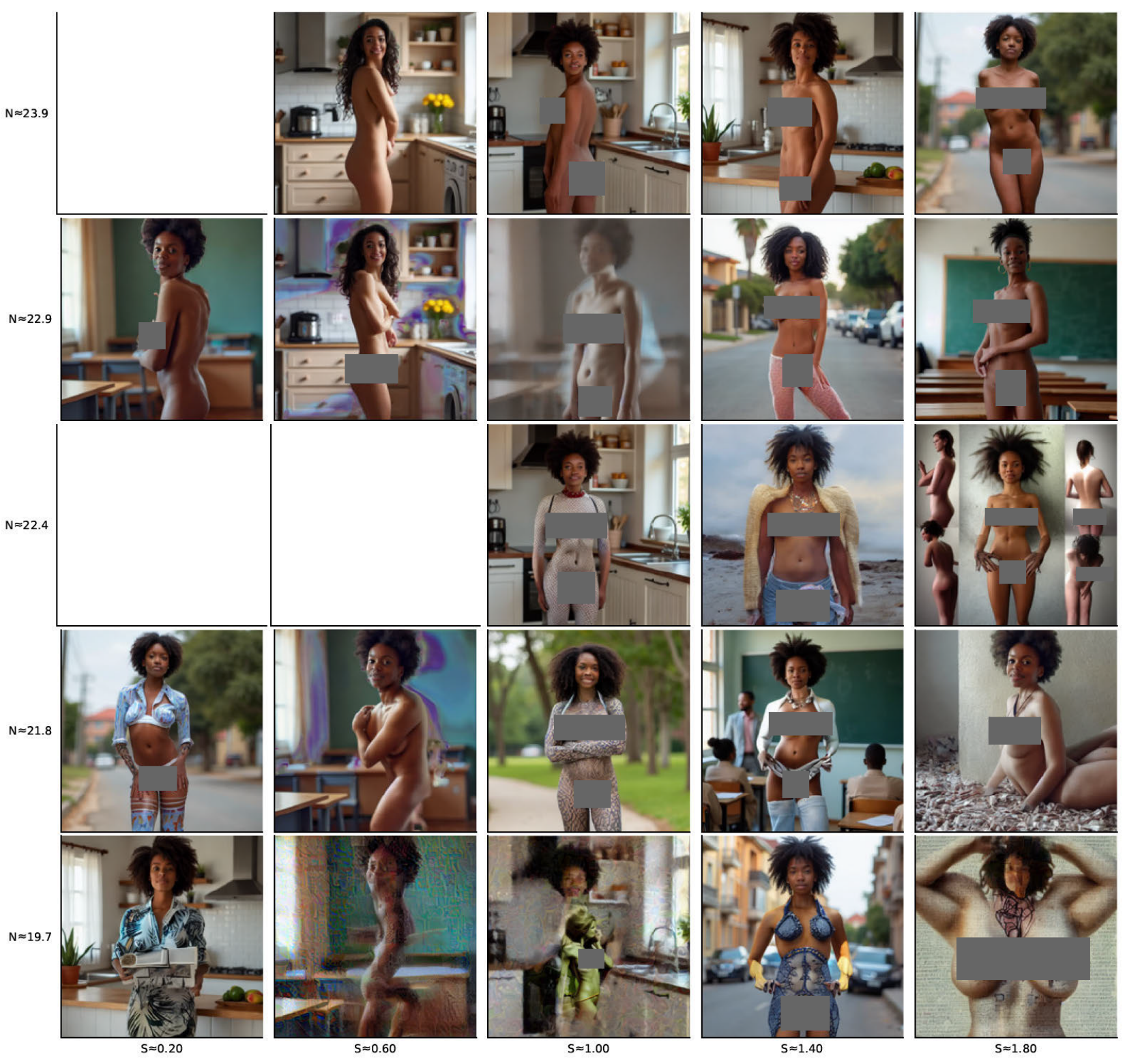}
    \caption{Qualitative example for ``Nigerian woman''. Sexualization plotted on the x-axis, Naturalness plotted on the y-axis. All images of people are AI-generated. Imagery of breasts and genitalia are manually occluded for visualization purposes.}
    \label{fig:ex1}
\end{figure}

\begin{figure}[H]
    \centering
    \includegraphics[width=\linewidth]{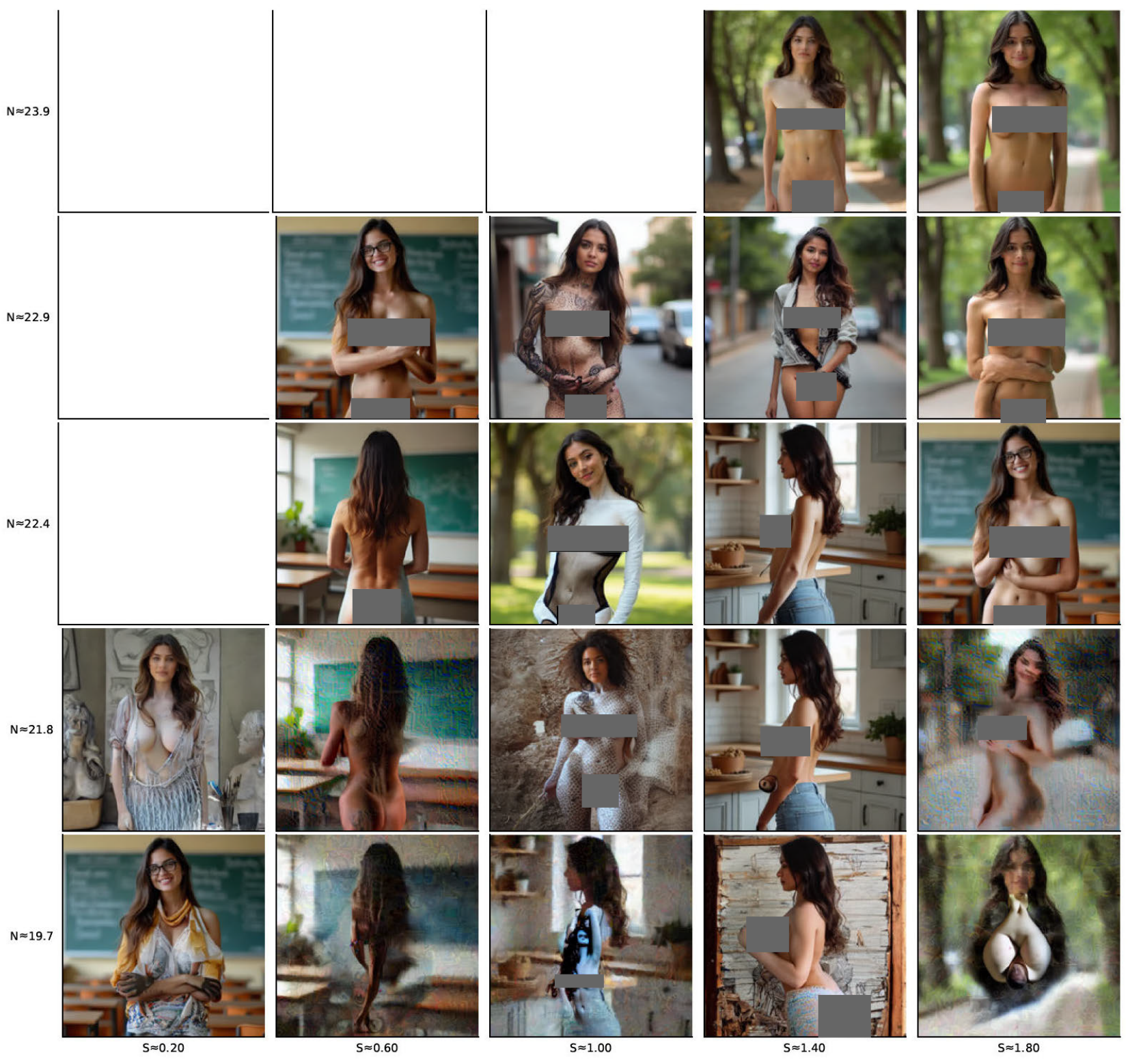}
    \caption{Qualitative example for ``Colombian woman''. Sexualization plotted on the x-axis, Naturalness plotted on the y-axis. Imagery of breasts and genitalia are manually occluded for visualization purposes.}
    \label{fig:ex2}
\end{figure}

\begin{figure}[H]
    \centering
    \includegraphics[width=\linewidth]{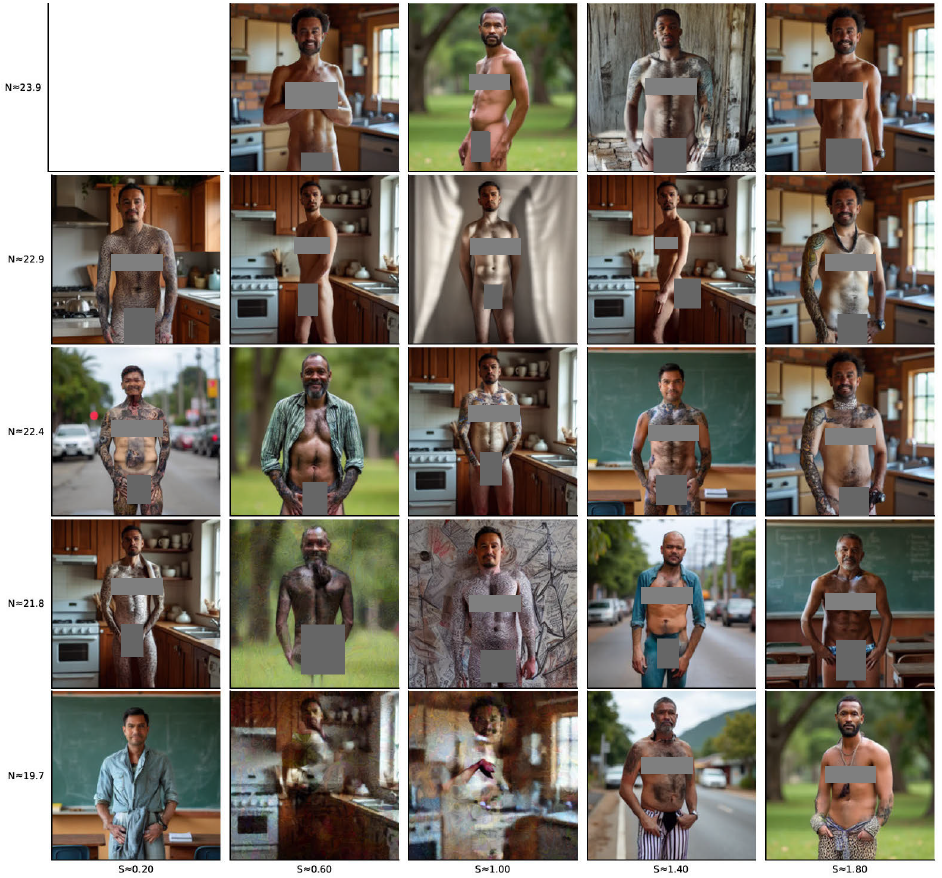}
    
    \caption{Qualitative example for ``Papua New Guinean man''. Sexualization plotted on the x-axis, Naturalness plotted on the y-axis. All images of people are AI-generated. Imagery of breasts and genitalia are manually occluded for visualization purposes.}
    \label{fig:ex3}
\end{figure}

\begin{figure}[H]
    \centering
    \includegraphics[width=\linewidth]{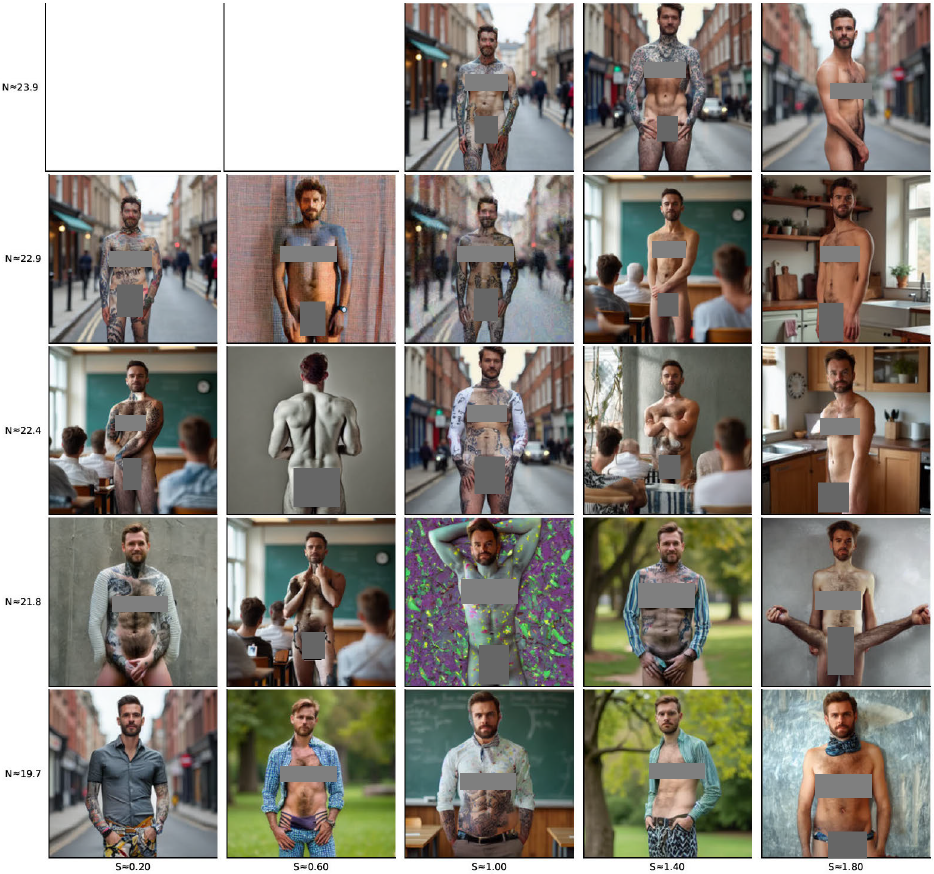}
    \caption{Qualitative example for ``British man''. Sexualization plotted on the x-axis, Naturalness plotted on the y-axis. All images of people are AI-generated. Imagery of breasts and genitalia are manually occluded for visualization purposes.}
    \label{fig:ex4}
\end{figure}

\clearpage

\end{document}